\documentclass[10pt,twocolumn,letterpaper]{article}

\usepackage{iccv}
\makeatletter
\@namedef{ver@everyshi.sty}{}
\makeatother
\usepackage{tikz}
\usepackage{times}
\usepackage{epsfig}
\usepackage{graphicx}
\usepackage{amsmath}
\usepackage{amssymb}
\usepackage{booktabs}
\usepackage{multirow,multicol}
\DeclareTextCommand{\textdschemical}{PU}{\9046\227}
\usepackage{todonotes}
\usepackage{color, colortbl}

\usepackage[pagebackref=true,breaklinks=true,letterpaper=true,colorlinks,bookmarks=false]{hyperref}

\iccvfinalcopy 

\ificcvfinal\fi

\begin{document}




\title{Fcaformer: Forward Cross Attention in Hybrid Vision Transformer}


\author{Haokui Zhang$^{\dag \ddag}$, Wenze Hu$^{\dag}$, Xiaoyu Wang$^{\dag}$\\  
  $^{\dag}$Intellifusion, Shenzhen, China\\
  $^{\ddag}$Harbin Institute of Technology (Shenzhen), Shenzhen, China\\
}

\maketitle
\ificcvfinal\thispagestyle{empty}\fi

\begin{abstract}

Currently, one main research line in designing a more efficient vision transformer is reducing the computational cost of self attention modules by adopting sparse attention or using local attention windows. In contrast, we propose a different approach that aims to improve the performance of transformer-based architectures by densifying the attention pattern. Specifically, we proposed forward cross attention for hybrid vision transformer (FcaFormer), where tokens from previous blocks in the same stage are secondary used. To achieve this, the FcaFormer leverages two innovative components: learnable scale factors (LSFs) and a token merge and enhancement module (TME). The LSFs enable efficient processing of cross tokens, while the TME generates representative cross tokens. By integrating these components, the proposed FcaFormer enhances the interactions of tokens across blocks with potentially different semantics, and encourages more information flows to the lower levels. Based on the forward cross attention (Fca), we have designed a series of FcaFormer models that achieve the best trade-off between model size, computational cost, memory cost, and accuracy. For example, without the need for knowledge distillation to strengthen training, our FcaFormer achieves 83.1\% top-1 accuracy on Imagenet with only 16.3 million parameters and about 3.6 billion MACs. 
This saves almost half of the parameters and a few computational costs while achieving 0.7\% higher accuracy compared to distilled EfficientFormer. 
\end{abstract}

\section{Introduction}
\label{sec:intro}

With the rapid adoption of transformer structures in the computer vision community, several types of attention patterns have been proposed to enhance the performance or speed of transformer models. For instance, ViT~\cite{dosovitskiy2020image} employs the vanilla global multi-head self-attention, Swin Transformers~\cite{liu2021swin} uses local windowed attention, MaxViT~\cite{tu2022maxvit} incorporates grid attention across interleaved tokens, and Dynamic ViT~\cite{rao2021dynamicvit} utilizes attention on progressively pruned tokens. These approaches aim to sparsify the attention patterns of the original ViT to achieve a better trade-off between speed and accuracy.

In contrast, we propose a new model block as well as a family of models called FcaFormer, which improves the performance of vision transformers by further densifying the attention patterns at a limited extra cost. Specifically, we propose to connect the input of the standard multi-head attention (MHA) module with extra tokens transformed from previous blocks in the same stage, while still restricting the attention module to output the original amount of tokens. To further reduce the computational cost, we merge the tokens from previous blocks by using depthwise convolutions with large strides. These tokens are further calibrated by scaling them with learned parameters, before being taken into the attention units in subsequent blocks.

The new forward cross attention connection has several advantages: 1) it helps transformers further exploit the interactions of tokens across different levels; 2) it reuses the previously generated tokens so that some of the information no longer needs to be preserved by the subsequent transformer operations, leading to potentially smaller models with similar accuracy; 3) similar to the residual connections in ResNet, this extra cross layer connection encourages more information flows to the lower levels of the network, which further accelerate the convergence. 

The newly densified connections come with a limited increase in computational cost. As explained in Section \ref{sec:computational_complexity}, this cost increase is linear rather than quadratic, since we keep the number of output tokens the same as in standard ViTs. Furthermore, most of the computation cost in most hybrid vision transformer architectures is in the feed forward network (FFN) rather than the MHA part of transformer blocks. Thus, the linear growth of computational complexity from densified connections does not significantly affect the overall computation cost. Finally, to further reduce the number of extra inputs, we use depthwise convolutions with large kernels and long strides to aggregate tokens from previous blocks.

We have incorporated the proposed Fca design into two typical classes of transformer models: the plain ViT model used in DeiT, and the hybrid ConvNet and transformer structures frequently seen in recent works \cite{mehta2021mobilevit,graham2021levit,li2022efficientformer}. Our experiments demonstrate that the Fca block can seamlessly replace the corresponding transformer blocks in these architectures, leading to significantly improved performance compared to their corresponding baselines. Specifically, FcaFormer-L1 achieves a top-1 accuracy of 80.3\% with approximately 6.2 million parameters and about 1.4 billion MACs. This is achieved while saving almost half the number of parameters, and achieving 1.1\% higher accuracy compared to the recently proposed EfficientFormer. Table \ref{tab:comparison} displays the comparison results.

The contribution of this paper is summarized as follows.

\begin{itemize}    
\setlength{\itemsep}{0pt}
    \item Opposite to many recent works that use sparse attentions to improve transformer models, we propose to design more efficient models by densifying the attention connection patterns, which open up a new and worthwhile research avenue for consideration. 
    
    \item We propose the FcaFormer block, which leverages existing tokens and enhance interactions across different levels. To achieve this, we introduce two new components: learnable scale factors (LSFs) and a token merge and enhancement module (TME). The LSFs allow us to effectively process cross tokens, while the TME generates representative cross tokens. Together, these components improve the performance of the FcaFormer models.  
    
    \item Based on the proposed FcaFormer block, we constructed several new models which have demonstrated better performance than various other recently proposed models.
    
\end{itemize}

\begin{figure*}[!ht]
\centering
\includegraphics[width=6.8in]{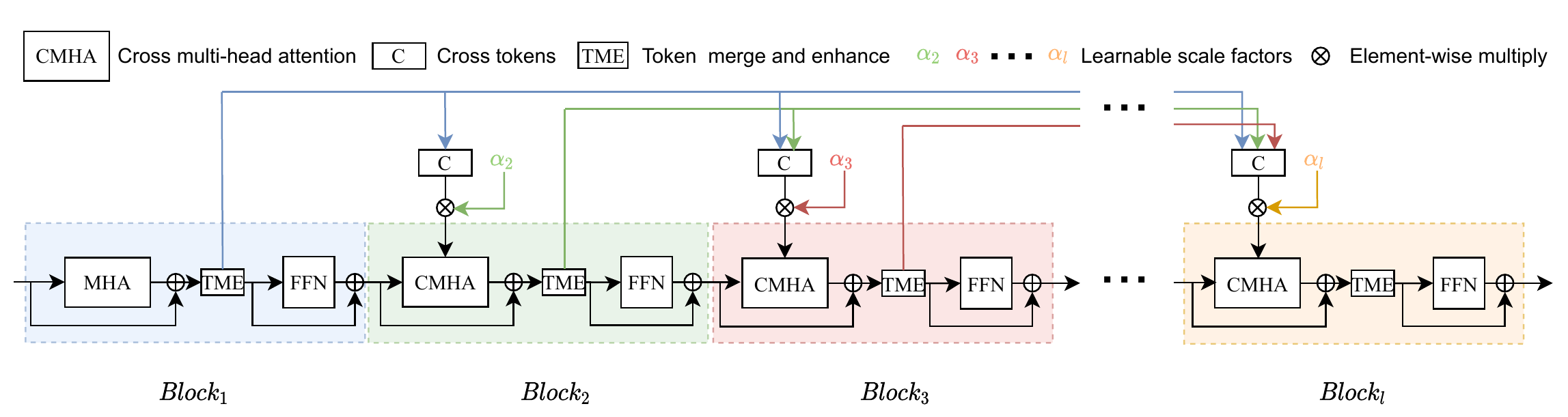}
\caption{A FcaFormer stage which consists of $L$ FcaFormer blocks. Compared with standard ViT blocks, we add token merge and enhancement (TME) part, which uses long stride large kernel convolutions to merge tokens spatially as cross tokens, and small kernel convolutions to further enhance tokens for channel mixing (FFN). The cross tokens are then used in later blocks as extra tokens for multi-head attention, after being calibrated by multiplying them with learned scaling factors (LSFs, $\alpha$).}
\label{fig:FcaFormer}
\end{figure*}

\section{Related Works}\label{sec::review}

\subsection{Pure vision transformers} \label{sec::pure_vit}

Dosovitskiy {\it et al.} introduced transformer model into vision tasks and proposed the ViT~\cite{dosovitskiy2020image}. It cropped an image into $16\times 16$ patches as an input token sequence to the transformer and used positional encoding to model spatial relations among tokens. DeiT~\cite{touvron2021training} lowered the difficulty of ViT model training by knowledge distillation, and achieved competitive accuracy with less pretraining data. To Further improve the model architecture, researchers attempted to optimize the ViT toward improving its computational efficiency. Among them, the Swin transformer~\cite{liu2021swin} computes self attention among shifted local windows. The MaxViT ~\cite{tu2022maxvit} uses block attention and grid attention alternatively to keep spatially global information exchange while significantly reduce the number of tokens involved in the self attention computation. The DynamicViT~\cite{rao2021dynamicvit} prunes redundant tokens progressively and dynamically depending on the input features. Mohsen {\it et al.}~\cite{fayyaz2022adaptive} proposed a differentiable parameter-free adaptive token sampler and plugged it into ViTs to sample part tokens for attention computation. 

Except for the DeiT, all methods above keep pure vision transformer architectures and seek to achieve better accuracy speed trade-off by using sparse attention via reducing the number of tokens in attention patterns. In contrast, we propose to achieve this goal by densifying the attention pattern, reusing existing tokens from previous blocks. Such interactions promote attentions across features of different semantic levels, which is very common in ConvNets and many methods before the wide adoption of deep learning.

\subsection{Hybrid ConvNet and vision transformers}

Rather than simplifying ViTs, another popular line of research is to combine elements of ViTs and ConvNets to form new backbones. Two early attempts can be found in \cite{graham2021levit, xiao2021early}, where ConvNet blocks are employed to extract low level information in early stages and ViT blocks are adopted in deep stages. Such a hybrid structure improves optimization stability and model performance. Similarly, BoTNet~\cite{srinivas2021bottleneck} replaces the standard convolution with multi-head attention in the last few blocks of ResNet. In~\cite{mehta2021mobilevit}, the deeper stages of MobileNetv2~\cite{sandler2018mobilenetv2} are replaced with their proposed MobileViT block. There are other hybrid models which mix convolution operation with self attention and channel mixer operations. For example, ConViT\cite{d2021convit} incorporates soft convolutional inductive biases via a gated positional self-attention. CMT~\cite{guo2022cmt} and Next-ViT~\cite{li2022next} insert both convolution operataion and self attention module into a single block. PVT v1~\cite{wang2021pyramid}, PVT v2~\cite{wang2022pvt}, LIT~\cite{pan2022less} and LIT v2~\cite{pan2022fast} insert convolutional operations into each stage of ViT models to reduce the number of tokens, and build hybrid multi-stage structures. 

Generally speaking, hybrid models achieve better trade-off between model cost and accuracy compared with pure vision transformer models. Therefore, we mainly focus on applying forward cross attention design on hybrid models to evaluate its effects. Our work is complementary to these hybrid design attempts, and can be used to replace the transformer part for most of these models.

\subsection{Skip connections in ConvNets}
\label{sec:skip_connect}

In retrospect, our work is also related to several key improvements in ConvNets design that introduces extra connections to improve the information flows. 

ResNet \cite{he2016deep} employs shortcut connections to overcome the degradation problem, where accuracy gets saturated and then degrades rapidly with the increasing convolutional network depth.
DenseNet \cite{huang2017densely} connects each layer to every other layer in a feedforward fashion. As with ResNet that builds the whole network by stacking several residual units, DenseNet consists of multiple dense blocks. Wang {\it et al.}~\cite{wang2018pelee} further improve DenseNet by using two-way dense layers to obtain different receptive fields. With limited extra computational cost, these model design choices solved bottlenecks existed in ConvNets and are still widely used in both academia and industry. 

Our proposed forward cross attention is similar to the works above in that it reuses existing intermediate results and introduces extra connections to the overall network structure. As experiments show, our work introduces similar benefits to transformer models such as better model performance and faster model convergence.

\subsection{Cross attention transformers}

In transformers, cross attention is usually used to mix two different embedding sequences. In~\cite{vaswani2017attention, gheini2021strengths}, the output of encoder is fed to decoder via cross attention. CrossViT~\cite{chen2021crossvit} mixes small-patch and large-patch tokens with cross attention to extract multi-scale feature. Our proposed FcaFormer shares the basic idea of integrating information with cross attention. However, unlike previous models, our FcaFormer integrates tokens from different semantic levels using TME and LSFs to overcome their distinct characteristics.

\section{The proposed FcaFormer}

\subsection{FcaFormer block and FcaFormer stage}

\label{sec: FcaFormer}

Fig.\ref{fig:FcaFormer} shows the major parts and connection relations of a FcaFormer stage which consists of $L$ FcaFormer blocks. Each FcaFormer block is composed of three major parts, which are cross multi-head attention (CMHA), token merge and enhancement (TME), and feed forward network (FFN) respectively. 

\textbf{CMHA and LSFs}. Different from the standard ViT block, our proposed FcaFormer block receives two sets of tokens as input. The block in turn generates two sets of tokens as well, which are denoted as $x^l$ and $\bar{x}^l$ respectively. Taking the block at depth $l$ as an example, the inputs are the regular tokens from its previous block $x^{l-1}$ and a set of block cross tokens $(\bar{x}^{l-2}, \bar{x}^{l-3} ... \bar{x}^1)$ from earlier blocks in the stage. \textit{The CMHA takes both $x^{l-1}$ and $(\bar{x}^{l-2}, \bar{x}^{l-3} ... \bar{x}^1)$ as input, but only generates $n$ tokens $\tilde{y^k}$ as its output}. 

It is worth noting that the cross tokens are scaled by  learned calibration coefficients (learnable scale factors LSFs) $\alpha$ before they are used in the CMHA. We found that the statistics of tokens from different semantic levels are very different. Without this calibration operation, cross tokens are hard to integrate to regular tokens in current blocks to work as we expected. Details are shown and analyzed in our ablation study part.

Specifically, given the inputs above, the query, key and value for multi-head attention modules are constructed as: 
\begin{eqnarray}
Q &=& W^Q x^{l-1} \\
K &=& W^K\left[x^{l-1}, (\mathbf{1} \alpha^{l})^T \otimes(\bar{x}^{l-2}, \cdots, \bar{x}^{1} ) \right] \\
V &=& W^V\left[x^{l-1},  (\mathbf{1} \alpha^{l})^T \otimes(\bar{x}^{l-2}, \cdots, \bar{x}^{1} ) \right]
\end{eqnarray}
where $W^K,W^V,W^Q$ are the three weight matrices transforming input into key, value and query tokens respectively. Elements in  $\alpha \in \mathcal{R}^{d \times 1}$ are the learnable scale factors for cross tokens and $\otimes$ denotes the element-wise multiplication where each token in $\bar{x}$ is multiplied by a corresponding scaling scalar in $\alpha$.

The computed $Q,K,V$ are then connected to the standard dot product attention operators as in the original transformer \cite{vaswani2017attention}, which computes the globally mixed intermediate tokens $y^l$. 

\begin{equation}\label{eqn::mha}
    y^l = x^{l-1}+W^P\left[\mathrm{softmax}\left(\frac{QK^T}{\sqrt{d}}+B\right)V\right]
\end{equation}
The $B$ in Eqn.\ref{eqn::mha} is a learnable matrix consisting of two parts, which are relative position bias for $x$ and relative depth bias for $\bar{x}$. The query has only $n$ tokens, so the output $y^l$ is of size $n \times d$. 

\begin{figure}[]
\centering
\includegraphics[width=2.6in]{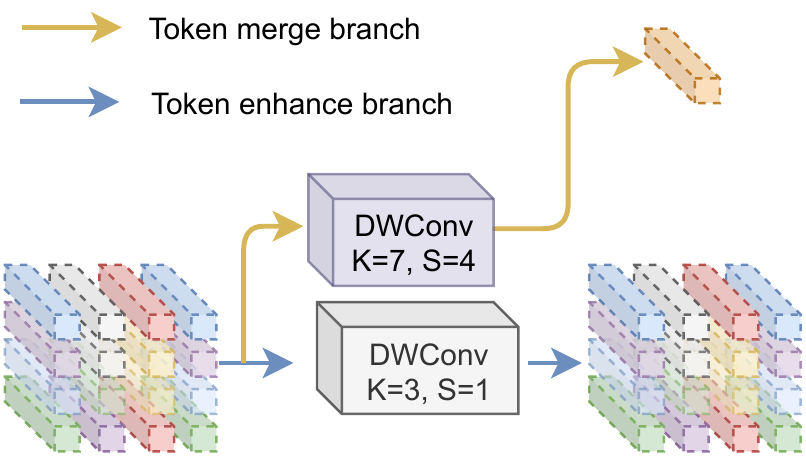}
\caption{The detailed structure of the token merge and enhancement (TME) module.}
\label{fig:TME}
\end{figure}

\textbf{TME}. The intermediate tokens $y^l$ are then passed to the added token merge and enhancement (TME) part, which computes the cross tokens $\bar{x}^l$ and locally enhanced tokens $z^l$ using two separate depthwise convolutions. As shown in Fig~\ref{fig:TME}, our TME has two branches. In the token merge branch, we use a large kernel $(7\times7)$ large stride $(s=4)$ depthwise convolution to generate $\bar{x}^l$, resulting in a small number of cross tokens to be used by subsequent blocks. In token enhancement branch, a standard depth wise convolution ($3\times3, s = 1)$ is used to locally mix the tokens so as to enhance the 2d spatial relations in the token sequence.

\textbf{FFN}. The locally enhanced tokens $z^l$ are then used in the standard FFN part of ViT to compute the output token $x^l$, which together with $\bar{x}^l$ are the entire output of a FcaFormer block at depth $l$. 

In summary, our FcaFormer are different from vanilla ViT in following ways:
\begin{itemize}   
\setlength{\itemsep}{0pt}
    \item \textbf{Asymmetric input and output}. In our FcaFormer, CMHA part takes both regular tokens and cross tokens as input but keep the output sequence length fixed to $n$ as in regular transformers. This is the key point that why our proposed FcaFormer only introduce limited extra computational cost. More details are explained in section~\ref{sec:computational_complexity}.     
   
    \item \textbf{Cross token scale}. LSFs are used to facilitate the integration of cross tokens into regular tokens. Without this calibration process, cross tokens may not function as expected. As shown in Fig~\ref{fig:learnable_sacle}.   
         
    \item \textbf{Relative depth embedding}. In the CMHA part, we choose to use relative positional encoding instead of absolute ones. Apart from its generally better performance, the choice also gives model the flexibility to encode relative depth of the cross tokens, which simplifies the model design. 
\end{itemize}

\begin{figure}[]
\centering
\includegraphics[width=\linewidth]{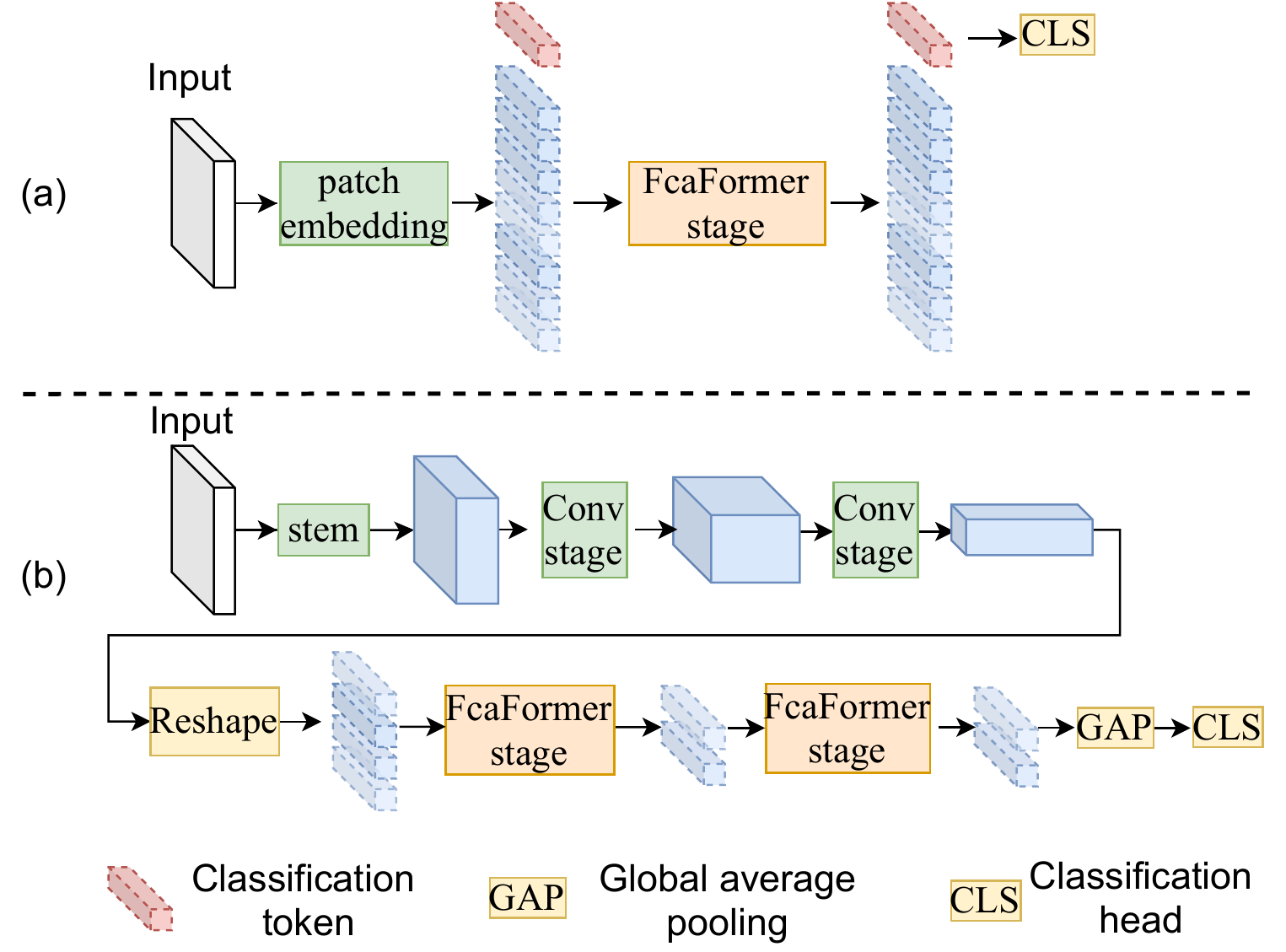}
\caption{The overall structure of FcaFormer Mobels for image classification tasks. (a) Plain FcaFormer model. This model is directly modified from the DeiT structure. (b) Hybrid FcaFormer model. The ConvNet stages are composed of ConvNext blocks. The detailed model scaling hyperparameters are specified in Sec.\ref{sec::model}}
\label{fig:FcaFormer_models}
\end{figure}

\subsection{FcaFormer Models}\label{sec::model}

The FcaFormer block is a generic block that can be grouped as FcaFormer stages to construct models. To demonstrate the universal effectiveness of our new attention pattern, we build two types of FcaFormer based models, each falls into one of the major categories of transformer related computer vision models as described in Sec. \ref{sec::review}. The overall structure of these models for classification task are illustrated in Fig.\ref{fig:FcaFormer_models}.

\textbf{Plain FcaFormer}. As shown in Fig~\ref{fig:FcaFormer_models} (a), following the style of vanilla ViT, we construct our plain FcaFormer model (denoted as FcaFormer-L0). The model essentially uses one block of FcaFormer after the patch embedding layer that crops $16 \times 16$ patches and converts them into a token sequence.  The corresponding task related head is kept unchanged compared with the original ViT model.

\textbf{Hybrid FcaFormer}. As illustrated in Fig~\ref{fig:FcaFormer_models} (b). Following the trend of combining ConvNet structures and transformer structures to build hybrid models, we also propose to build hybrid FcaFormer. Following LeViT~\cite{graham2021levit} and ViT-C\cite{xiao2021early}, we adopt the most straightforward way to build our Hybrid FcaFormers, where there are two conventional ConvNet stages followed by two FcaFormer stages. The ConvNet stage is composed of ConvNext blocks~\cite{liu2022convnet} plus a downsampling layer that uses pointwise and depthwise convolutions to reduce the feature map resolution by 4$\times$ and increases the feature map channels. 

Based on the above overall structure, we build models of different sizes to compare with other works. The key scaling hyper-parameters is summarized below, where $D$ denotes the channel size:

\begin{itemize}
    \item FcaFormer-L0: D=192, L=12
    \item FcaFormer-L1: D=(64,128,192,320), L=(2,2,6,2)
    \item FcaFormer-L2: D=(96,192,320,480), L=(2,2,7,2)
\end{itemize}

\subsection{Computational Complexity of FcaFormer}
\label{sec:computational_complexity}

\begin{figure}
\includegraphics[width=3.0in]{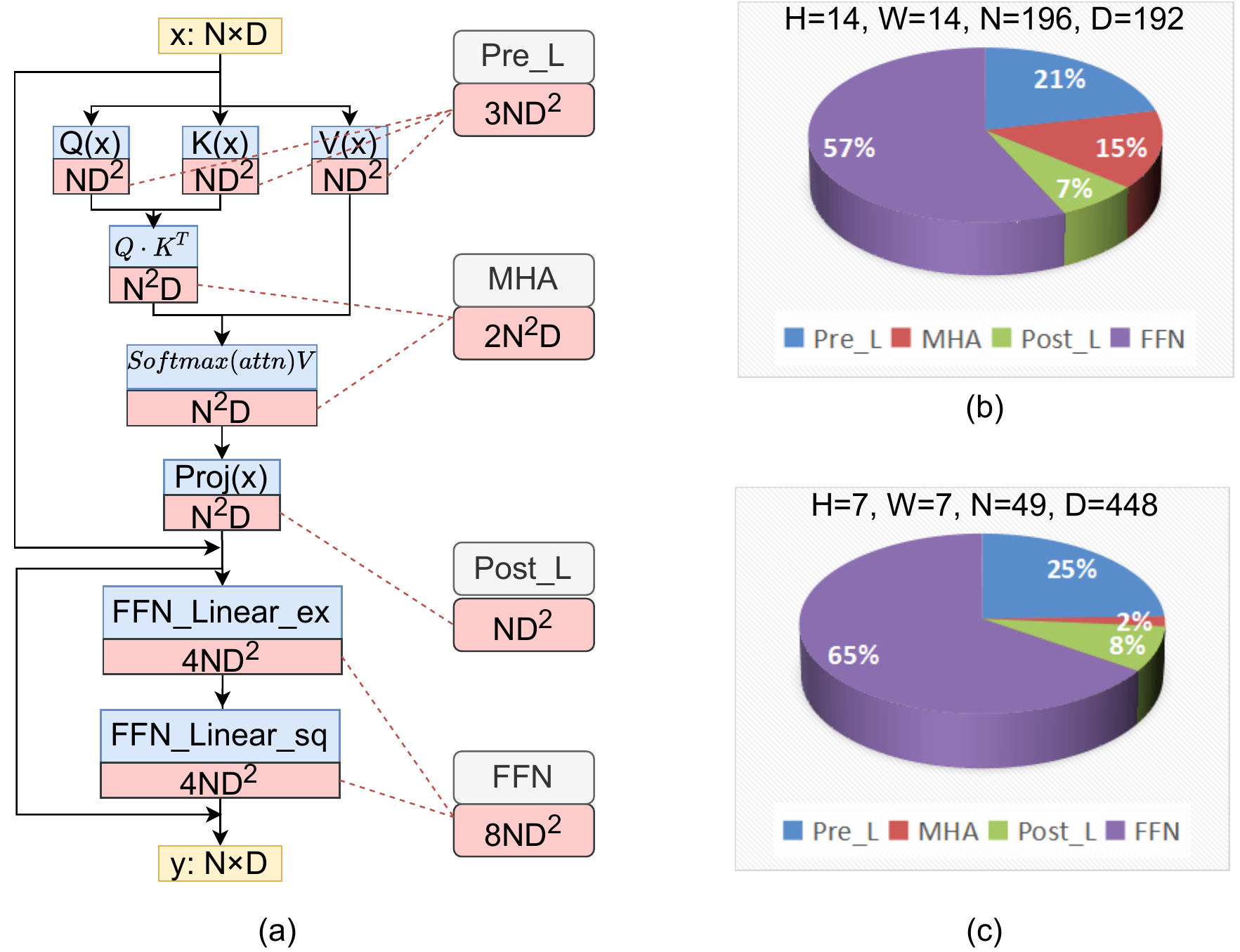}
\caption{Computational cost analysis. (a) Detailed analysis on the number of parameters and MACs for a transformer block; (b)-(c) Computation decompositions in blocks of stage 3 of DeiT-Tiny and stage 4 of EfficientFormer-L1 respectively.}
\label{fig::FLOPs}
\end{figure}

In terms of computational complexity, transformers are notorious for its $\mathcal{O}(n^2)$ scaling property, which is why so many prior works try to reduce the inputs to the MHA model. However, the extra computational cost of densifying the attention pattern is rather limited in our FcaFormer models, because of the following reasons:

1) The added extra computation is restricted to the MHA part, since we keep the output sequence length of MHA fixed to $n$ as in regular transformers. This leads to linear scaling cost in terms of the number of additional input tokens, which itself is a small ( $(l-1)*n/s^2$ ) compared with $n$. The computational cost of FFN part is kept unchanged compared with original ViT blocks. 

2) In vision transformer model families, the FFN part actually constitutes the majority of computations. This is paradoxical to the fact that MHA scales at $\mathcal{O}(n^2)$  while FFN scales at $\mathcal{O}(n)$, largely because that the neglected constant token size $d$ in big O analysis is rather large and scales poorly.  Fig. \ref{fig::FLOPs} details the exact adds and multiplications (MACs) of a standard ViT block in terms of the number of tokens $n$ and the dimension of tokens $d$. Taking the transformer block DeiT-Tiny as an example, because $n=196$ and $d=192$ are at the same scale, the $2n^2d$ MACs in MHA is much less compared with that ($8nd^2$) of FFN (15\% vs 56\% respectively). For models in pyramid shapes such as swin transformers where $n < d$, the majority skews even more towards FFN because of the local windowing effect. Compared with the reference models, our FcaFormer variants only introduce about 13\% extra computational cost. 

It looks like that the proposed FcaFormer is heavier than its reference model as it introduces extra computations. However, the counter-intuitive fact is that \textit{FcaFormer can be more light-weight than its reference model}. As is mentioned in introduction, FcaFormer reuses tokens from previous blocks, enhances interaction of tokens across blocks and improve information flow. Compared with the vanilla ViT, FcaFormer has better parameter efficiency. Therefore, FcaFormer can get comparable or even better performance with fewer layers and channels. Our experiments further verifies this advantage.

In addition to the detailed analysis above, we want to point out that the prior efforts to sparsify the attention patterns spatially in Sec.\ref{sec::pure_vit} and our proposal to densify the attention patterns across semantic levels are complementary to each other. While not the key emphasis of this paper, it is definitely worth exploring if the two types of designs can be combined to yield even more efficient and performant models.

\begin{table}[]
\centering
\begin{tabular}{lccc}
\toprule
Models         &   \# params. &  FLOPs &  Top1 acc  \\
               &   (M)        &  (G)    &   (\%)      \\
\hline
DeiT-T~\cite{touvron2021training}         &    5.5       &         &  72.2           \\
\rowcolor{lightgray} 
FcaFormer-L0    &    5.9.      &         &  74.3           \\
\hline
Swin-1G$^{\ddag}$        &    6.3      &  1.5    &  78.4          \\   
\rowcolor{lightgray} 
FcaFormer-L1    &    6.2       &  1.4    &  80.3          \\
\hline
ConvNext-Tiny~\cite{liu2022convnet}   &    29       &  4.5    &  82.1          \\
Swin-Tiny~\cite{liu2021swin}         &    29       &  4.5    &  81.3          \\   
\rowcolor{lightgray} 
FcaFormer-L2     &    16.3      &  3.6    &  83.1        \\
\bottomrule 
\end{tabular}
\caption{Comparison with baselines. The plain version FcaFormer is based on DeiT-T. The hybrid ViT FcaFormer models are based on ConvNext and Swin Transformers. \ddag indicate our inplementation}
\label{tab:baselines}
\end{table}

\section{Experiments}
\label{sec::experiments}

In this section, we conduct image classification experiments on Imagenet-1K~\cite{deng2009imagenet}, semantic segmentation experiments on ADE20K~\cite{zhou2017scene}, and object detection experiments on MS-COCO~\cite{2014Microsoft} to evaluate our proposed models. We first compare the proposed FcaFormers with our baselines and the previous SOTA methods. Then, we conduct detailed study to show the effectiveness of our design choices. 

\begin{table}
\setlength\tabcolsep{2pt}
\centering
\begin{tabular}{lclccccc}
\toprule
Models             & KD   &  Type    & Param. & MACs & Top1 \\
\hline 
DeiT-Tiny~\cite{touvron2021training}          & Y    &  ViT     & 6         &  -        & 74.5        \\
DeiT-Tiny~\cite{touvron2021training}          & -    &  ViT     & 6         &  -        & 72.2        \\
\rowcolor{lightgray} 
FcaFormer-L0        & -    &  ViT     & 6         &  -        &  74.3       \\
\hline
LeViT-128~\cite{graham2021levit}          & Y    &  Hybrid  & 9         &  0.4      &  78.6       \\          
EfficientFormer-L1~\cite{li2022efficientformer} & Y    &  Hybrid  & 12          &  1.3      &  79.2       \\
ParCNet~\cite{zhang2022parc}            & -    &  Conv    & 5         &  1.7      &  78.6       \\
TNT-Ti~\cite{han2021transformer}           & -     &  ViT    & 6         &  1.4      &  73.9     \\
Swin-1G$^\dag$~\cite{liu2021swin}            &  -    &  ViT    & 7         &  1.0      &  77.3      \\
Swin-2G$^\dag$~\cite{liu2021swin}            &  -    &  ViT    & 13        &  2.0      &  79.2      \\
EfficientFormer-L1$^{\ddag}$ & -    &  Hybrid & 12          &  1.3      &  
76.1       \\
MobileViT-V1~\cite{mehta2021mobilevit}       & -    &  Hybrid & 6         &  2.0      &  78.4       \\
EdgeViT-XS~\cite{pan2022edgevits}         & -    &  Hybrid  & 7         &  1.1      &  77.5       \\
MobileViTV2\cite{sandler2018mobilenetv2}        & -    &  Hybrid  & 5         &  1.8      &  78.1       \\ 
CoaT-Tiny~\cite{xu2021co}          & -    &  Hybrid  & 6         &  4.4      &  78.3       \\
PVT-V2-B1~\cite{wang2022pvt}             & -    &  Hybrid  & 13        &  2.1      &  78.7       \\ 
Mobile-Former~\cite{chen2022mobile}      &  -    & Hybrid  & 14        &  0.5      &  79.3       \\
Edgenext~\cite{maaz2023edgenext}           &  -    & Hybrid  & 6         &  1.3      &  
79.4       \\
MobileOne-S4~\cite{vasu2022improved}          &  -    & Hybrid  & 15        &  3.0      & 79.4 \\
\rowcolor{lightgray} 
FcaFormer-L1        &  -    &  Hybrid &  6     &  1.4     &  \textbf{80.3}  \\
\hline
DeiT-S~\cite{touvron2021training}             &  Y    &  ViT    &  22       &  4.6     &  81.2  \\
LeViT-256~\cite{graham2021levit}          &  Y    &  Hybrid &  19     &  1.1     &  81.6  \\
EfficientFormer-L3~\cite{touvron2021training} &  Y    &  Hybrid &  31     &  3.9     &  82.4  \\
ResNet50~\cite{he2016deep}           &  -    &  Conv   &  25       &  4.1     &  78.8  \\
ResNet50$^{\ddag}$          &  -    &  Conv   &  25       &  4.1     &  79.1  \\
PoolFormer-S24~\cite{yu2022metaformer}     &  -    &  Conv   &  21     &  3.4     &  80.3  \\
PoolFormer-S36~\cite{yu2022metaformer}     &  -    &  Conv   &  31     &  5.0     &  81.4  \\
ConvNext-Tiny~\cite{liu2022convnet}         &  -    &  Conv   &  29       &  4.5     &  82.1  \\
VAN-B2~\cite{guo2022visual}             &  -    &  Conv   &  27       &  5.0     &  82.8  \\
DeiT-S~\cite{touvron2021training}             &  -    &  ViT    &  22       &  4.6     &  79.9 \\
Swin-T~\cite{liu2021swin}             &  -    &  ViT    &  29       &  4.5     &  81.3  \\ 
T2T-ViT-14~\cite{yuan2021tokens}         &  -    &  ViT    &  22       &  4.8     &  81.5  \\
T2T-ViT-19~\cite{yuan2021tokens}         &  -    &  ViT    &  39       &  8.5     &  81.9  \\
MViTv2-T~\cite{li2022mvitv2}           &  -    &  ViT    &  24       &  4.7     &  82.3  \\
CSWin-T~\cite{dong2022cswin}            &  -    &  ViT    &  23       &  4.3     &  82.7  \\
MobileViTV2~\cite{sandler2018mobilenetv2}        &  -    &  Hybrid &  19     &  7.5     &  81.2  \\
LITV2~\cite{pan2022fast}              &  -    &  Hybrid &  28       &  3.7     &  82.0  \\
Next-ViT-S~\cite{li2022next}         &  -    &  Hybrid &  32     &  5.8     &  82.5  \\
\rowcolor{lightgray} 
FcaFormer-L2        &  -    &  Hybrid &  16     &  3.6     & \textbf{83.1} \\
\hline
\end{tabular}
\caption{Comparison with the state-of-the-art methods on ImageNet-1K validation set. KD means knowledge distillation is used during training. $\ddag$ indicates implemented by us, where models are trained following the training setting used in ConvNext. Accuracy and FLOPs are calculated on input image size 224$\times$ 224. $^\dag$ borrowed from~\cite{chen2022mobile}}
\label{tab:comparison}
\end{table}

\subsection{ImageNet classification}

\textbf{Experiment settings}. The FcaFormer-L0 is implemented based on code of DeiT~\footnote{https://github.com/facebookresearch/deit}. The hybrid FcaFormer models are implemented based on code of ConvNext~\footnote{https://github.com/facebookresearch/ConvNeXt} and Swin~\footnote{https://github.com/microsoft/Swin-Transformer}.  We follow the training recipes in DeiT \cite{touvron2021training} to train our FcaFormer-L0, except that we did not use the knowledge distillation. To train the hybrid FcaFormers, we use the same training hyper parameters and augmentations as used in ConvNext except that the batch size is restricted to 1024 and the initial learning rate is reduced to 2e-3. This change is because we don't have enough GPUs to support the default batch size of 4096 in ConvNext.

\textbf{Comparison with baselines}.
The results are summarized in Table~\ref{tab:baselines}, from which we can see that both types of our proposed FcaFormer models outperform their reference models by a significant margin. Compared with DeiT-T, FcaFormer-L0 does have 0.4 M more parameters, but it achieves 2.1\% higher accuracy. Our hybrid FcaFormers achieve better performance in all three metrics: accuracy, model size and computational cost. FcaFormer-L1 achieves 1.9\% higher top-1 accuracy compared with similarly sized Swin-1G. FcaFormer-L1 also has good scalability. The scaled up version FcaFormer-L2 outperforms the ConvNext-Tiny by 1.0\% with 20\% less computational cost and 44\% fewer parameters. FcaFormer-L2 also surpasses Swin-T by 1.8\% while using fewer parameters and less computation. 

In summary, our proposed forward cross attention design universally improves performances of both plain version and hybrid ViT models without increasing too much extra computation cost or even saving parameters and FLOPs. 

\textbf{Comparison with other models}. In Table~\ref{tab:comparison}, we make a comparison with other models proposed in recent two years. Compared to the latest state-of-the-arts, including ConvNets, ViTs and Hybrid models, our proposed FcaFormer models achieve the best accuracy under the condition of having similar model sizes. In addition, the proposed models beat models strengthened by knowledge distillation in classification accuracy.  

Specifically, FcaFormer-L1 outperforms MobileOne and EdgeNext by 0.9\% in classification accuracy, while keeping comparable or smaller model size and fewer computational cost. Among models having about 25 million parameters, FcaFormer-L2 achieves the highest classification accuracy with the fewest parameters. Compared with Next-ViT-S which achieves the second highest accuracy, FcaFormer-L2 saves about half parameters and 38\% computational cost, while gaining 0.6 percentage points higher accuracy. Also, note that both EfficientFormer-L3 and LeViT-256 are trained with knowledge distillation, which has been verified to improve accuracy significantly~\cite{touvron2021training}. Even so, FcaFormer-L2 still has better accuracy and smaller model size compared with these two models.

\subsection{Semantic segmentation}

\begin{table}
\setlength\tabcolsep{3pt}
\centering
\begin{tabular}{llccccc}
\toprule
Method    &   Backbone  & mIOU  & \# params. & MACs \\
\hline
DANet     &  ResNet-101 & 45.2  &  69  &  1119  \\
DpLab.v3+ &  ResNet-101 & 44.1  &  63  &  1021  \\ 
ACNet     &  ResNet-101 & 45.9  &  -   &  -     \\
DNL       &  ResNet-101 & 46.0  &  69  &  1249  \\
OCRNet    &  ResNet-101 & 45.3  &  56  &  923   \\
UperNet   &  ResNet-101 & 44.9  &  86  &  1029  \\
UperNet   &  DeiT III (ViT-S)   & 46.8  &  42  & 588  \\
\hline
UperNet   &  Swin-T     & 45.8  &  60  & 945  \\

UperNet   &  ConvNext-T & 46.7  &  60  & 939  \\
\rowcolor{lightgray} 
UperNet   & FcaFormer-L2 & 47.6  &  46  & 730 \\
\hline 
\end{tabular}
\caption{Semantic segmentation on the ADE20K dataset. We use UperNet as our segmentation method and compare our performance using the same method with other popular backbones.}
\label{tab:seg}
\end{table}

\textbf{Training setting}. To evaluate the proposed FcaFormers on downstream tasks. We apply them on the ADE20K semantic segmentation task.  Following Swin and ConvNext, we adopt UperNet~\cite{xiao2018unified} as our base framework and implement segmentation experiments on mmseg~\footnote{https://github.com/open-mmlab/mmsegmentation}. FcaFormer-L2 is trained for 160K iterations with a batch size of 16. Model pretrained on ImageNet-1K is used to initialize segmentation model. More details are presented in supplementary materials.

\textbf{Comparison with baselines}. The results are summarized in Tab.\ref{tab:seg}, compared with Swin-Tiny, FcaFormer-L2 achieves 1.8\% higher mIoU. Compared with ConvNext-Tiny, FcaFormer-L2 achieves 0.9\% higher mIoU.  Compared with Swin-T and ConvNext-T, the last two stages of FcaFormer-L2 are narrower (384 vs 320, 768 vs 480), which saves parameters and computational cost. Therefore, our FcaFormer-L2 achieves higher mIOU, while saving 23\% parameters and 22-23\% computation cost. 

\subsection{Object detection}

\textbf{Training setting}. Object detection experiments are conducted on COCO 2017 and implemented on mmdet~\footnote{https://github.com/open-mmlab/mmdetection}. Following \cite{liu2021swin} and \cite{liu2022convnet}, we fintune MASK-RCNN and Cascade Mask R-CNN on the COCO dataset with FcaFormer pretrained on ImageNet-1k. We use multi-scale training, AdamW optimizer, and a 3× schedule. More details are presented in supplementary materials.

\textbf{Comparison with baselines}. The results are summarized in Tab.\ref{tab:det}. Experimental results on objection detection show a similar trend with that on semantic segmentation. Compared with Swin-Tiny and ConvNext-Tiny, FcaFormer-L2 achieves higher values on all six metrics, while having fewer parameters and less computational cost. 

\begin{table}[t]
\renewcommand\arraystretch{1.2}
\setlength\tabcolsep{0.1pt}
\centering
\scalebox{0.9}{
\begin{tabular}{lcccccccc}
\toprule
Backbone & \#Params. & MACs & $AP^{b}$ & $AP^{b}_{50}$ & $AP^{b}_{75}$ &$AP^{m}$ & $AP^{m}_{50}$ & $AP^{m}_{75}$\\
\hline 
\multicolumn{9}{c}{Mask-RCNN 3$\times$ schedule} \\
SWin-T  &  48 &  267  & 46.0 & 68.1 & 50.3 & 41.6 & 65.1 & 44.9  \\
ConvNext-T  & 48  &  262  & 46.2 & 67.9 & 50.8 & 41.7 & 65.0 & 44.9 \\
\rowcolor{lightgray} 
FcaFormer-L2 & 37  &  249  & 47.0 & 68.9 & 51.8 & 42.1 & 65.7 & 45.4 \\ 
\hline
\multicolumn{9}{c}{Cascade Mask-RCNN 3$\times$ schedule} \\
ResNet-50   & 82  & 739  & 46.3 & 64.3 & 50.5 & 40.1 & 61.7 & 43.4  \\
DeiT-S      & 80  & 889  & 48.0 & 67.2 & 51.7 & 41.4 & 64.2 & 44.3  \\
X101-32     & 101 & 819  & 48.1 & 66.5 & 52.4 & 41.6 & 63.9 & 45.2  \\
X101-64     & 140 & 972  & 48.3 & 66.4 & 52.3 & 41.7 & 64.0 & 45.1  \\ 
Swin-T      & 86  &  745  & 50.4 & 69.2 & 54.7 & 43.7 & 66.6 & 47.3 \\
ConvNext-T  & 86   &  741  & 50.4 & 69.1  & 54.8 & 43.7 & 66.5 & 47.3 \\
\rowcolor{lightgray}
FcaFormer-L2 & 74  &  728   & 51.0 & 69.4  & 55.5 & 43.9 & 67.0 & 47.4 \\
\hline 
\end{tabular}
}
\caption{Object detection on the COCO dataset.}
\label{tab:det}
\end{table}

\subsection{Ablation study}

\begin{figure}
\centering
\includegraphics[width=2.6in]{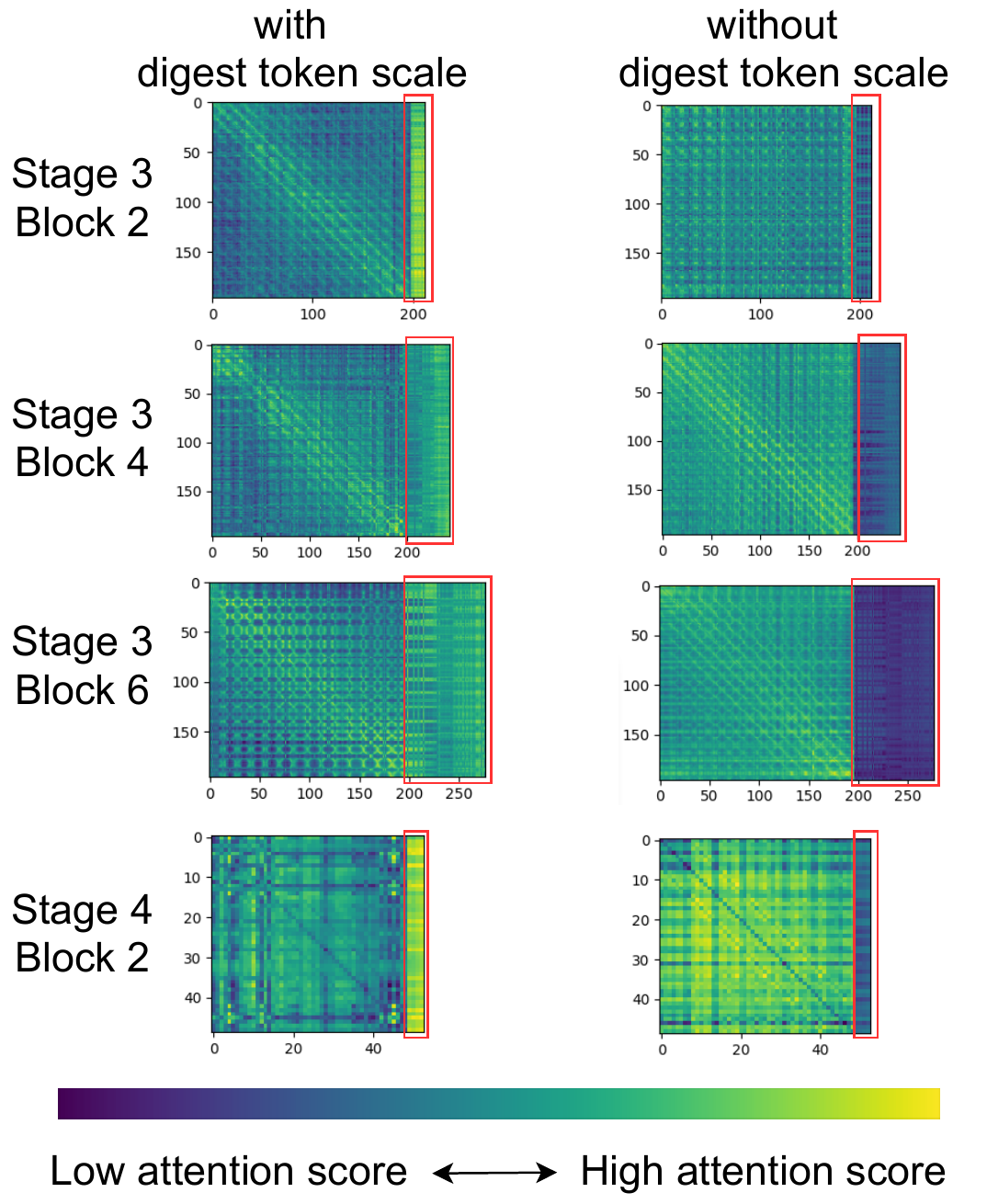}
\caption{Effects of LSFs. Attention scores for cross tokens are marked with red boxes. From rows 1-4, attention maps are from blocks 2,4 and 6 of stage3, and block 2 of stage 4 respectively. }
\label{fig:learnable_sacle}
\end{figure}

\begin{table*}
\setlength\tabcolsep{3pt}
\centering
\begin{tabular}{cllccccl}

\toprule
Rows  & Models        & Model differences     & \# Param. (M)  & MACs (B)  & Latency (ms) & Memory (M) & Top1 acc (\%)  \\
\hline
1   &  Swin-1G       &  baseline 
                               &  6.25      &    1.49   &  340   &  46.64   & 78.4   \\   
2   &  ConvSwin      & +early convolution 
                               &  6.11      &    1.26   &  269   &  38.34   & 78.8 (+0.4)  \\
 \rowcolor{gray!20}                               
3   &  ConvViT       & +global attention
                               &  6.11      &    1.32   &  300   &  42.01   & 79.4 (+1.0)  \\
 \rowcolor{gray!40}
4   &  FcaFormer     & +naive forward cross attention.
                               &  6.11      &    1.37   &  311   &  42.07   & 79.4 (+1.0)  \\ 
 \rowcolor{gray!40}
5   &  FcaFormer    & +learnable scale factors. 
                               &  6.11      &  1.37     &  311   &  42.07   & 79.9 (+1.5)  \\                              
\rowcolor{gray!60} 
6   &  FcaFormer    & +TME.
                               &  6.19      &  1.37     &  312   &  42.31   & 80.3 (+1.9)  \\
\rowcolor{gray!80} 
7   &  FcaFormer    & scale up to L2
                               &  16.3      &  3.6     &  728   &  95   & 83.1   \\
                            
\bottomrule
\end{tabular}
\caption{Ablation study on Imagenet-1K.  Steps 1-7 depict the process we followed to develop our FcaFormer models from the Swin-1G. We evaluated the latency and memory usage of the models on ARM Quad Core Cortex-A17. To conduct our experiments, we utilized the RK3288 platform, which is commonly employed in real-world applications such as smart TV and AI entrance guard systems. }
\label{tab:ablation_components}
\end{table*}

In this section, we conduct ablation analysis on components proposed in our FcaFormer. 



\textbf{Effects of the learnable scale factors (LSFs)}. Fig~\ref{fig:learnable_sacle} shows the attention score maps of several FcaFormer blocks trained with and without LSFs. It can be seen that without LSFs, the attention between cross tokens and regular tokens are generally weak, and become weaker as the depth goes deepr ( down on the figure). This verifies our hypothesis that characteristics of tokens in different levels can be very different. Comparing the two columns of attention maps, it clearly shows that the LSFs helps increasing the correlation between regular tokens and the cross tokens, thus encourages the reuse of previously generated tokens.

\textbf{From baseline to FcaFormer}. The high performance of FcaFormer models is based on two key factors: combining the strengths of ViTs and ConvNets, and utilizing a dense but not heavy forward cross attention design. Table~\ref{tab:ablation_components} shows how we integrate these two key points in designing our models. To evaluate the performance of the proposed FcaFormer on real-world applications, we deployed each model on the edge device Rockchip 3288, which is widely used in various embedded applications. We collected latency and memory usage information for each model.

The first row in Table~\ref{tab:ablation_components} provides a baseline micro model to speed up experiments. The second row uses the simplest way to build a hybrid model, which inherits some advantages of ConvNets and ViTs. Compared to the baseline, the early convolution structure achieves higher accuracy while requiring less computation and memory usage. In the third row, we replaced window attention with global attention based on computational complexity analysis in section~\ref{sec:computational_complexity}. This improved the accuracy by 0.6\% while introducing very limited extra cost. However, latency increased by 11\%, and memory usage increased by 7.7 M.

\begin{table}[t]
\setlength{\abovecaptionskip}{0.1cm}
\setlength{\belowcaptionskip}{-0.1cm}
\renewcommand{\arraystretch}{1.0}
\setlength\tabcolsep{0.1pt}
\centering
\scalebox{0.95}{
\begin{tabular}{lcccccccc}
\toprule
Models         &  \# params. &  MACs & Latency  & Memory & Acc  \\
               &   (M)        &  (B)  & ARM(ms)  & (M)  &    (\%) \\ 
\hline
ConvNext-Tiny     &    29       &  4.5   & 875    & 129  &  82.1  \\
ConvNext-Small    &    50       &  8.7   & 1618   & 211  &  83.1  \\
ConvNext-Base     &    89       &  15.4  & 2708   & 364  &  83.8  \\
ConvNext-Large    &    198      &  34.4  & 5604   & 764  &  84.3  \\
\hline
Swin-Tiny     &    29       &  4.5   & 855    & 139  &  81.3  \\
Swin-Small    &    50       &  8.7   & 1576   & 222  &  83.0  \\
Swin-Base     &    88       &  15.4  & 2624   & 378  &  83.5  \\
\hline
FcaFormer-L1(Micro)   &    6.2      &  1.4   & 312   & 42   &  80.3  \\ 
FcaFormer-L2(Tiny)    &    16       &  3.6   & 728   & 95   &  83.1  \\ 
FcaFormer-L3(Small)   &    28       &  6.7   & 1344  & 148  &  84.2  \\
FcaFormer-L4(Base)    &    66       &  14.5  &  2624 & 328  &  84.9  \\
\bottomrule 
\end{tabular}
}
\caption{Batch size=1, image size=224, four threads. {\bf ARM:}Quad Core Cortex-A17.}
\label{tab:rk3288}
\end{table}

Rows 4 and 5 attempt to introduce forward cross attention. In row 4, cross tokens are not calibrated, which only introduces extra cost and has no benefit to accuracy. In row 5, we adopted LSFs to adjust cross tokens, which further improved the accuracy by 0.5\%. So far, FcaFormer has outperformed the baseline by 1.5\% while utilizing fewer parameters and incurring lower computational costs compared to the baseline.

In row 6, we introduced the TME module to generate representative cross tokens and enhance regular tokens, which improved the final accuracy to 80.3\%, 1.9\% higher than the baseline. Finally, we scaled up the micro size model and got FcaFormer-L2. This new model achieved top-1 accuracy of 83.1\%, meeting the requirements for some practical applications with low latency, taking less than 1 second and minimal memory usage, below 100M.

\subsection{Test on different devices}

\begin{figure}
  \centering
  \centering
  \includegraphics[width=0.49\linewidth]{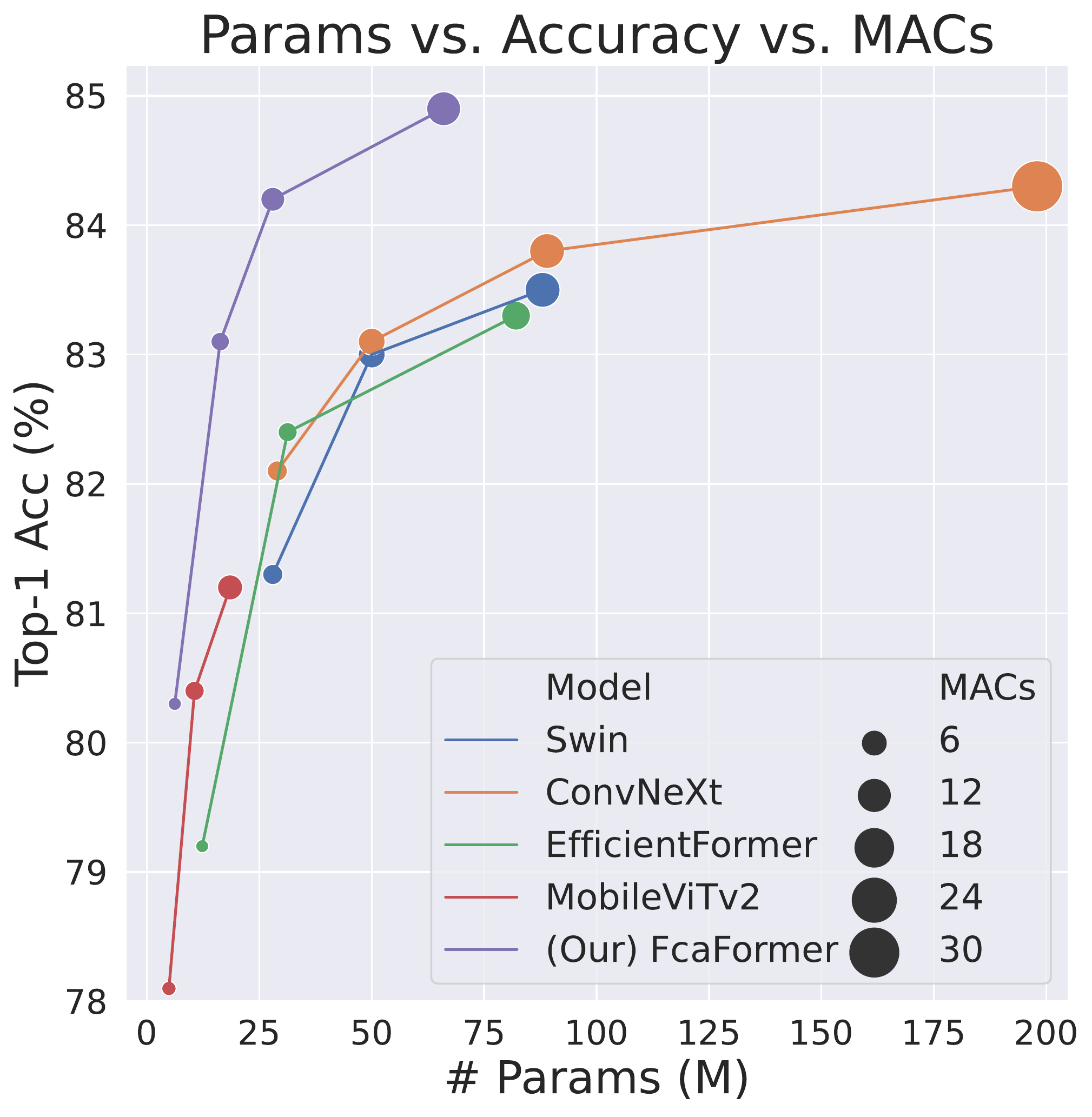}
  \includegraphics[width=0.49\linewidth]{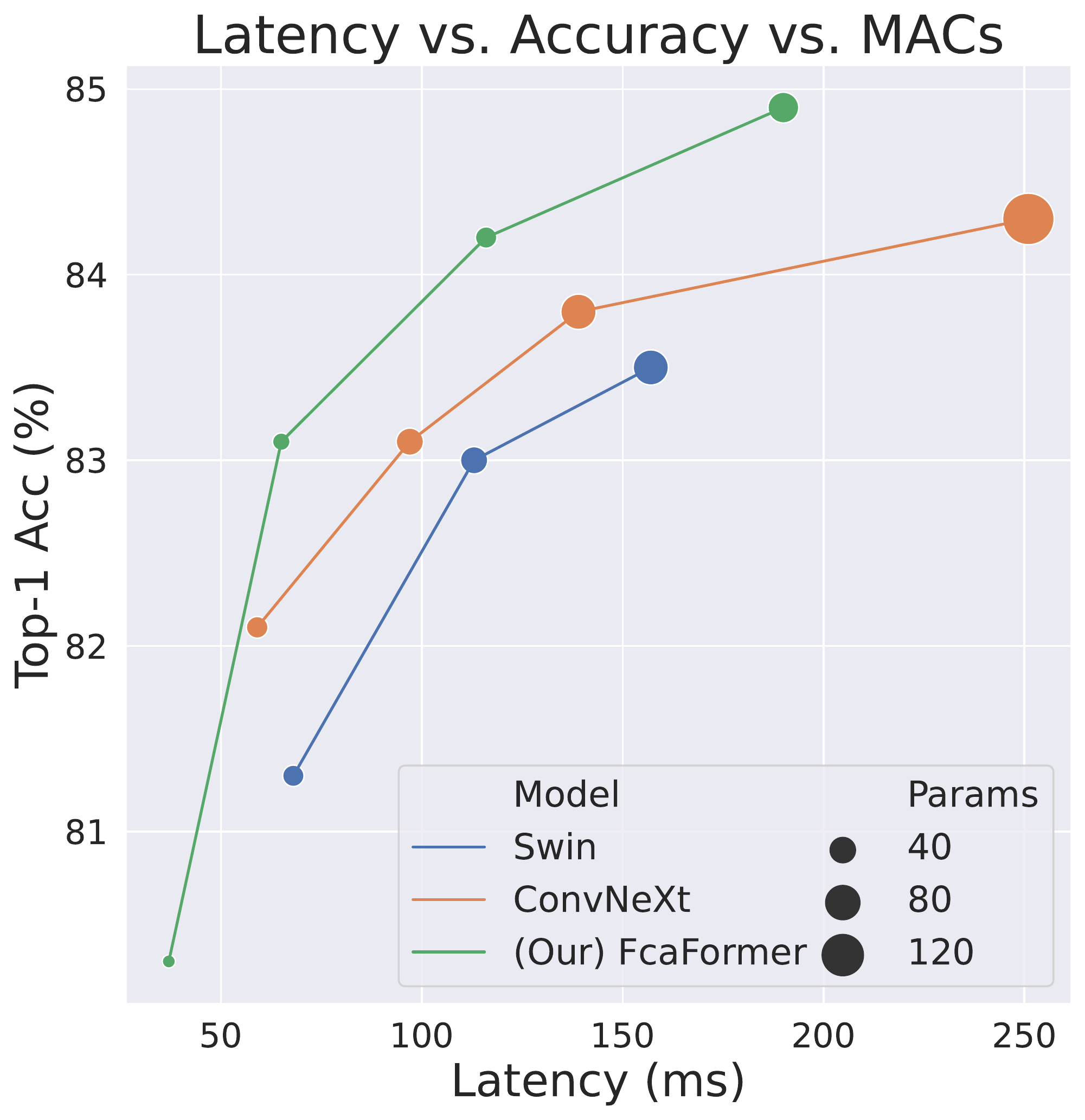}
  \caption{Comparison experiments. Left: Params vs. acc vs. MACs. Right: Latency vs. acc vs. MACs. The latency is measured on a single \textbf{NVIDIA RTX 3090} GPU with batchsize=64.}
  \label{fig:trade_offs}
  \vspace{-3mm}
\end{figure}

We deployed our proposed models on two different devices, the widely used edge device RK3288 and GPU device RTX3090, to test their inference efficiency. We also built larger models, FcaFormer-L3 (D=(96,192,320,512), L=(3,6,12,3)), and FcaFormer-L4 (D=(128,256,512,768), L=(3,6,12,3)), to validate the scalability of FcaFormer. We repeated two sets of experiments for 100 and 1000 times, respectively. The average cost is listed in Table~\ref{tab:rk3288} and shown in Fig~\ref{fig:trade_offs}. Our models achieved higher accuracy compared with ConvNexts while having far fewer parameters, less memory usage, and lower latency. Furthermore, our network demonstrated good scalability. From tiny model to base model, FcaFormers consistently maintained a clear advantage compared to both Swin and ConvNext models.

\section{Discussions}

In this paper, we introduce a new type of attention pattern for hybrid vision models. This attention pattern leverages previously generated tokens to create forward cross-attentions that span different semantic levels. Our experiments show that this approach is effective across different model scales and vision tasks. For future work, we propose combining this design with prior research on sparsifying spatial attention patterns, which could lead to even more efficient model backbones for a range of applications

{\small
\bibliographystyle{ieee_fullname}
\bibliography{egbib}
}

\end{document}